\documentclass{article}

\usepackage{microtype}
\usepackage{graphicx}
\usepackage{subcaption}
\usepackage{booktabs}

\usepackage{hyperref}

\usepackage[accepted]{icml2026_cats}

\usepackage{amsmath}
\usepackage{amssymb}
\usepackage{mathtools}
\usepackage{amsthm}
\usepackage{pifont}
\usepackage{mdframed}

\usepackage[capitalize,noabbrev]{cleveref}

\theoremstyle{plain}

\theoremstyle{definition}

\theoremstyle{remark}

\icmltitlerunning{Knowledge Updating via Memory Dynamics}

\begin{document}

\twocolumn[
  \icmltitle{Continual Knowledge Updating in LLM Systems:\\
    Learning Through Multi-Timescale Memory Dynamics}

  \icmlsetsymbol{equal}{*}

  \begin{icmlauthorlist}
    \icmlauthor{Andreas Pattichis}{cyi}
    \icmlauthor{Constantine Dovrolis}{cyi,gt}
  \end{icmlauthorlist}

  \icmlaffiliation{cyi}{The Cyprus Institute, Nicosia, Cyprus}
  \icmlaffiliation{gt}{School of Computer Science, Georgia Institute of Technology, Atlanta, GA, USA}

  \icmlcorrespondingauthor{Andreas Pattichis}{a.pattichis@cyi.ac.cy}
  \icmlcorrespondingauthor{Constantine Dovrolis}{c.dovrolis@cyi.ac.cy}

  \icmlkeywords{continual learning, large language models, memory-augmented LLMs, retrieval-augmented generation, synaptic consolidation, multi-timescale dynamics, selective forgetting}

  \vskip 0.3in
]

\printAffiliationsAndNotice{}

\begin{abstract}
LLMs are trained once, then deployed into a world that never stops changing. External memory compensates for this, but most systems manage it explicitly rather than letting it adapt on its own. Biological memory works differently: coupled multi-timescale dynamics make new associations immediately usable, strengthen what repetition confirms, and let the rest fade. We argue that external memory should follow a similar principle. In \textit{Memini}, this view takes the form of an associative memory that organizes knowledge as a directed graph. Each edge carries two coupled internal variables, one fast and one slow, following the Benna-Fusi model of synaptic consolidation. From this coupling, episodic sensitivity, gradual consolidation, and selective forgetting are expected to emerge as facets of a single mechanism, reframing external memory as a learning substrate that reorganizes through its own dynamics. This workshop article describes an early-stage conceptual design without experimental evaluation.
\end{abstract}

\section{Introduction} \label{sec:intro}

Large language models (LLMs) are increasingly deployed in settings where factual knowledge does not remain static \cite{dhingraTimeAwareLanguageModels2022, lazaridouMindGapAssessing2021}. New facts emerge, old ones become obsolete, and associations between concepts strengthen, weaken, or disappear. A model frozen at deployment cannot keep up with any of this. The challenge is not merely one of access to new information. It is one of selective adaptation: strengthening what ongoing evidence confirms, letting go of what it no longer supports, and doing so continuously as new experience arrives.

Existing work addresses this through two main paradigms, and neither resolves the problem at the level it requires. The first is parametric continual learning, which adapts the model by updating its weights \cite{shiContinualLearningLLM2025}. This changes the language processing substrate directly, but at significant cost: parameter updates risk catastrophic forgetting, are computationally expensive, and are often unavailable to downstream developers who access the model as a service \cite{delangeContinualLearningSurvey2022, sunBlackBoxTuningLanguageModelasaService2022}. These difficulties are compounded by a more fundamental limitation. Different kinds of knowledge change at different rates, yet a single parameter-update mechanism treats all evidence as equally worth absorbing. This makes forgetting both more likely and harder to control \cite{wuContinualLearningLarge2024}. The second paradigm keeps the backbone frozen and augments it with external memory \cite{mialonAugmentedLanguageModels2023, wolfeShiftModelsCompound}. This improves access to prior information, but the mechanisms that store, organize, and retrieve do not themselves change over time. The memory may grow or be pruned, but it does not reorganize: associations are not strengthened by repeated evidence, not weakened by absence, and retrieval does not shift as new information arrives.

We therefore argue for a sharper distinction. A system whose memory merely grows is accumulating. A system whose memory consolidates, forgets, and reshapes its organization in response to experience is learning \cite{kumaranWhatLearningSystems2016}. This distinction motivates a different paradigm, one in which learning happens through structured memory dynamics. \textbf{The central claim is that continual learning in LLM systems can be implemented as the reorganization of associative memory through multi-timescale dynamics.} Prior memory-augmented LLM systems store or retrieve information; they do not learn by reshaping their association structure over time. Three principles define this paradigm. First, dynamics: associations evolve through reinforcement and decay, not just by being appended to. Second, multi-timescale learning: fast processes capture recent evidence while slow processes consolidate what repetition confirms, and these coexist on every association \cite{bennaComputationalPrinciplesSynaptic2016}. Third, selectivity: the system does not store everything or retain everything, and what it forgets is as consequential as what it keeps \cite{norbyWhyForgetAdaptive2015}.

We instantiate this paradigm in \textit{Memini}\footnote{\textit{Memin\={\i}} is Latin for ``I remember'' or ``I hold in mind.'' Grammatically perfect in form but present in meaning, it denotes a present state of retention shaped by prior experience.}, a persistent directed association graph in which each edge carries two coupled internal variables, one fast and one slow, following the Benna-Fusi synaptic consolidation model \cite{bennaComputationalPrinciplesSynaptic2016}. The interaction between these two variables is intended to produce episodic sensitivity, gradual consolidation, and selective forgetting as emergent behaviors of the same dynamics, without separate stores, explicit rules, or gating modules. Retrieval operates through spreading activation over these evolving edge weights, so the same query issued at different times follows different paths and recovers different context. The novelty is not an external memory, a graph, or a retrieval procedure. It is that the memory itself is a dynamical state, driven directly by the incoming document stream rather than by retrieval activity or an external manager such as an agent or LLM revising it from outside. Its current organization reflects prior experience and shapes the assimilation of future experience.

\section{Related Work} \label{sec:background}

Continual learning for LLMs has been studied extensively as a parametric problem \cite{shiContinualLearningLLM2025, wuContinualLearningLarge2024}, with dominant approaches including replay, regularization, architectural expansion, and targeted knowledge editing. In all of these approaches, learning occurs in the model's parameters, and the central challenge is managing the stability-plasticity tradeoff so that new knowledge can be absorbed without overwriting what came before \cite{delangeContinualLearningSurvey2022}.

Other approaches keep the backbone frozen and augment it with external memory \cite{mialonAugmentedLanguageModels2023}. These range from vector-store retrieval-augmented generation \cite{lewisRetrievalAugmentedGenerationKnowledgeIntensive2020} to agent memory with tiered hierarchies \cite{packerMemGPTLLMsOperating2024} or memory streams with periodic reflection and summarization \cite{parkGenerativeAgentsInteractive2023}. But in each case, the management is external, applied by fixed rules rather than driven by ongoing evidence.

Within this paradigm, three lines of work push back against this rigidity, each in a different direction. The first adds structural organization to memory through entity graphs, PageRank retrieval, or LLM-revised note links \cite{edgeLocalGlobalGraph2025, gutierrezRAGMemoryNonParametric2025, xuAMEMAgenticMemory2025}, but the edges remain static as new documents arrive. The second makes retrieval adaptive over a graph through spreading activation or query-aware reweighting \cite{jiangSYNAPSEEmpoweringLLM2026, pavlovicLeveragingSpreadingActivation2025, lauBreakingStaticGraph2026}, yet the weights (fixed at indexing, set by time decay, or modulated only per query) are never shaped by repeated evidence. The third targets forgetting through single-timescale decay with recall-driven reinforcement \cite{zhongMemoryBankEnhancingLarge2024, hondaHumanLikeRememberingForgetting2026}, but a single timescale cannot be both fast enough for recent evidence and slow enough to preserve what repetition confirms.

Table~\ref{tab:comparison} summarizes the gap: no system combines evolving edge dynamics, multi-timescale consolidation, and emergent selective forgetting. The memory may be structured, traversed, or decayed, but it does not learn.

\begin{table}[t]
    \centering
    \caption{Memory-augmented LLM systems compared on the three properties that together define Memini: \emph{evolving} edge weights driven autonomously by incoming evidence, \emph{multi-timescale} dynamics coupling fast access with slow consolidation, and \emph{selective forgetting} of unsupported associations. Explicit deletion, per-query reweighting, threshold archival, and recall-driven reinforcement do not qualify.}
    \label{tab:comparison}
    \small
    \setlength{\tabcolsep}{2pt}
    \begin{tabular}{lccc}
        \toprule
        & \textbf{Evolving} & \textbf{Multi-} & \textbf{Selective} \\
        \textbf{System} & \textbf{weights} & \textbf{timescale} & \textbf{forgetting} \\
        \midrule
        Standard RAG    & \ding{55} & \ding{55} & \ding{55} \\
        MemGPT          & \ding{55} & \ding{55} & \ding{55} \\
        GraphRAG        & \ding{55} & \ding{55} & \ding{55} \\
        HippoRAG~2      & \ding{55} & \ding{55} & \ding{55} \\
        A-MEM           & \ding{55} & \ding{55} & \ding{55} \\
        SYNAPSE, SA-RAG, CatRAG  & \ding{55} & \ding{55} & \ding{55} \\
        MemoryBank, ACT-R-LLM & \ding{55} & \ding{55} & \ding{51} \\
        \midrule
        \textbf{Memini} & \ding{51} & \ding{51} & \ding{51} \\
        \bottomrule
    \end{tabular}
\end{table}

\section{System Design} \label{sec:memory}

\subsection{Memory Architecture} \label{sec:memory:architecture}
 
\begin{figure*}[t]
    \centering
    \includegraphics[width=\textwidth]{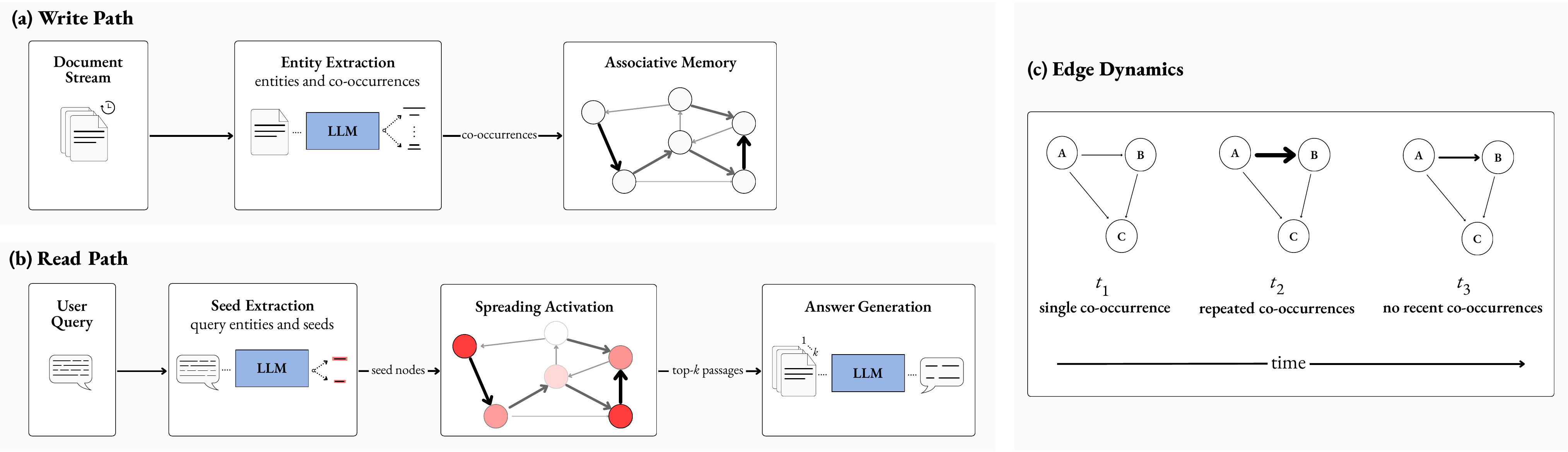}
    \caption{Overview of \textit{Memini}.
    \textbf{(a)~Write path.} The LLM extracts entities and co-occurrences from arriving documents. Each co-occurrence updates an edge of the association graph through coupled fast and slow variables. Edge thickness reflects $w_{\text{fast}}$.
    \textbf{(b)~Read path.} Entities from the user's query activate seed nodes (red). Activation propagates along $w_{\text{fast}}$-weighted edges for a bounded number of steps, and the passages associated with the highest-activation nodes are returned as context for answer generation.
    \textbf{(c)~Edge dynamics.} A single edge ($A \to B$) tracked across three moments while the rest of the graph is held fixed. A single co-occurrence creates a transient trace ($t_1$). Repeated co-occurrences consolidate the association ($t_2$). Without recent reinforcement it fades but persists ($t_3$).}    
    \label{fig:architecture}
\end{figure*}

The memory is a directed graph $G = (V, E)$ in which nodes represent entities or concepts identified in the incoming document stream and edges represent directed associations between them. Three structural properties distinguish these edges from those in prior graph-based systems (Table~\ref{tab:comparison}). They are unlabeled, with meaning encoded in the pattern, strength, and direction of connectivity rather than in explicit relation types. They are directed, with $A \to B$ distinct from $B \to A$, reflecting the asymmetry of associative strength found in spike-timing-dependent plasticity \cite{biSynapticModificationsCultured1998}. They carry persistent, evolving weights that change as new documents arrive.

Each directed edge $(A, B)$ carries an internal state consisting of two coupled variables, $w_{\text{fast}}$ and $w_{\text{slow}}$. Together, the tuple $(A, B, w_{\text{fast}}, w_{\text{slow}})$ constitutes the minimal unit of memory in this system. The fast variable $w_{\text{fast}}$ is the accessible component. It responds directly to co-occurrence events, decays quickly without reinforcement, and is the sole variable read during retrieval. The slow variable $w_{\text{slow}}$ is the hidden consolidation component. It receives no direct external input, changes only through bidirectional coupling with $w_{\text{fast}}$, and decays much more slowly. Although $w_{\text{slow}}$ is never read directly, it influences retrieval by sustaining $w_{\text{fast}}$ at non-zero levels for associations that have been consolidated through repeated reinforcement. The full state of the memory at any time $t$ is therefore the graph $G(t) = (V(t), E(t))$ together with the edge states $\{(w_{\text{fast}}(e, t),\, w_{\text{slow}}(e, t))\}_{e \in E(t)}$.

The backbone LLM and the memory system occupy different layers (Figure~\ref{fig:architecture}). The LLM reads and writes text; the memory does the learning. This separation has two consequences. The backbone can be replaced or upgraded without disturbing what the memory has learned, and the memory can evolve continuously without modifying the backbone.

\subsection{Multi-Timescale Edge Dynamics} \label{sec:memory:dynamics}

Without dynamics, the graph can only grow by appending new nodes and edges, and associations set at creation remain fixed indefinitely. A single-timescale update rule cannot resolve the resulting plasticity-stability tension. Learning fast enough to incorporate new evidence risks overwriting what came before, while learning slowly enough to retain prior structure prevents timely adaptation. Work on synaptic consolidation has shown that memory systems require multiple internal timescales to remain both plastic and stable over extended experience \cite{bennaComputationalPrinciplesSynaptic2016, zenkeTheoriesSynapticMemory2025}, a principle that generalizes the cascade model of \citet{fusiCascadeModelsSynaptically2005}. The same multi-timescale principle has been validated independently in optimization theory \cite{behrouzNestedLearningIllusion2025} and in empirical neural architectures \cite{behrouzTitansLearningMemorize2025}.

We adopt the simplest instantiation of the Benna-Fusi chain model, with two coupled variables per edge ($n = 2$) following \citet{kaplanisContinualReinforcementLearning2018}. The dynamics on each directed edge $(A, B)$ are:
\begin{equation}
\frac{d w_{\text{fast}}}{dt}
= -\frac{w_{\text{fast}}}{\tau_{\text{fast}}}
+ C\!\left(w_{\text{slow}} - w_{\text{fast}}\right)
+ I(t),
\label{eq:wfast}
\end{equation}
\begin{equation}
\frac{d w_{\text{slow}}}{dt}
= -\frac{w_{\text{slow}}}{\tau_{\text{slow}}}
+ C\!\left(w_{\text{fast}} - w_{\text{slow}}\right),
\label{eq:wslow}
\end{equation}
where $\tau_{\text{fast}}$ and $\tau_{\text{slow}}$ are the decay time constants ($\tau_{\text{slow}} \gg \tau_{\text{fast}}$), $C$ is the coupling strength, and $I(t)$ is the co-occurrence-driven input defined as $I(t) = b$ when concepts $A$ and $B$ co-occur in a document at time $t$, and $I(t) = 0$ otherwise. 

Three forces act on $w_{\text{fast}}$: decay toward zero at rate $1/\tau_{\text{fast}}$, coupling that pulls it toward $w_{\text{slow}}$, and input from co-occurrence events. Only two forces act on $w_{\text{slow}}$: a much slower decay, and coupling that pulls it toward $w_{\text{fast}}$. The key structural property is that $w_{\text{slow}}$ receives no direct external input. It accumulates only indirectly, through coupling, when $w_{\text{fast}}$ is repeatedly elevated by co-occurrence events, and once accumulated, it sustains $w_{\text{fast}}$ through the reverse direction of the same coupling. There is no explicit negative input when $A$ appears without $B$; weakening is handled through the decay terms, so that prolonged absence of reinforcement is functionally equivalent to gradual depression.

During retrieval, the system reads $w_{\text{fast}}$ alone. This variable reflects both recency, through direct boosts from recent co-occurrence events, and consolidation, through the sustained level provided by coupling from $w_{\text{slow}}$. If $w_{\text{slow}}$ were included in the retrieval weight, a recently mentioned association and a deeply consolidated one could appear similar, and the recency signal would be lost. Two edges with identical current $w_{\text{fast}}$ but different $w_{\text{slow}}$ values will diverge over subsequent time steps. The edge with high $w_{\text{slow}}$ remains retrievable as coupling sustains it, while the edge with low $w_{\text{slow}}$ decays. The system thereby differentiates episodic from consolidated associations through its own dynamics.

These dynamics make retention conditional on repeated evidence. Associations that the evidence stream continues to support consolidate and persist, while those that no longer receive reinforcement weaken and eventually disappear. This selective retention is not imposed by an explicit rule, a threshold, or a gating module, but emerges from the dynamics themselves. The role of forgetting is thereby reframed. Continual learning has conventionally treated it as uniformly harmful, a degradation to be minimized \cite{delangeContinualLearningSurvey2022}. In a system where knowledge changes over time, however, selective decay is what keeps the memory aligned with the current state of the world \cite{norbyWhyForgetAdaptive2015}. A system that retains every association indefinitely accumulates stale and contradictory evidence that degrades retrieval quality. Under this framing, what the system forgets becomes as consequential as what it retains.

\subsection{Emergent Memory Properties} \label{sec:memory:emergent}

Three characteristic behaviors are expected to follow from the dynamics without requiring separate stores, explicit rules, or additional modules. First, a single co-occurrence creates an immediately retrievable but transient episodic trace. $w_{\text{fast}}$ rises sharply but decays if no further evidence arrives, while $w_{\text{slow}}$ barely changes. Second, repeated co-occurrence produces consolidation. The coupling term gradually pulls $w_{\text{slow}}$ upward, and once $w_{\text{slow}}$ has accumulated sufficiently, it sustains $w_{\text{fast}}$ between reinforcement events, so that the association remains retrievable without depending on recent input. This transition from episodic trace to stable association parallels the episodic-to-semantic transition described by CLS theory \cite{mcclellandWhyThereAre1995} and Tulving's memory taxonomy \cite{tulvingPrecisElementsEpisodic1984}, but here it emerges from dynamics on a single structure rather than from explicit dual systems. Third, when a consolidated association stops being reinforced, both variables decay jointly but more slowly than either would alone. The coupled decay departs from the simple exponential profile of single-timescale systems and, in the general Benna-Fusi chain model, approaches the power-law form established empirically for human memory \cite{wixtedGenuinePowerCurves1997}.

These dynamics also compound. Each co-occurrence event lands on an edge whose current state reflects its entire prior history, so earlier consolidation shapes how later evidence is absorbed. The resulting memory state is trajectory-dependent rather than merely data-dependent, since the same set of documents arriving in a different order produces a different associative landscape.

\subsection{Retrieval via Spreading Activation} \label{sec:memory:retrieval}

Retrieval follows the same associative logic as the memory structure \cite{collinsSpreadingactivationTheorySemantic1975}. Entities mentioned in a query are matched to graph nodes and activated simultaneously as seed nodes, each receiving initial activation $u_i^{(0)} = 1$. Activation then propagates outward along directed, weighted edges according to
\begin{equation}
u_i^{(t+1)} = (1 - \delta)\, u_i^{(t)}
+ \sum_{j \in \mathcal{N}^{-}_{i}} S \cdot
\frac{w_{\text{fast}}(j \to i)}{\deg_{\text{out}}(j)}\, u_j^{(t)},
\label{eq:spreading}
\end{equation}
where $\delta$ is a per-iteration retention decay, $S$ is a global spreading factor, $w_{\text{fast}}(j \to i)$ is the persistent dynamic edge weight, and $1/\deg_{\text{out}}(j)$ is the fan effect that penalizes high-degree hub nodes \cite{andersonRetrievalPropositionalInformation1974}. This propagation equation is adapted from \citet{jiangSYNAPSEEmpoweringLLM2026}, with the key difference that $w_{\text{fast}}$ values here are persistent and history-shaped rather than computed per query. 

The process is a deterministic, constrained propagation procedure that runs for a fixed number of iterations $T$, respects edge direction, and terminates without stochastic sampling. After $T$ iterations, nodes are ranked by final activation score and the top-$k$ associated passages are retrieved. Retrieval reads the current memory state but does not modify it; $w_{\text{fast}}$ and $w_{\text{slow}}$ remain unchanged.

Because retrieval operates over experience-shaped weights, the system's effective retrieval behavior changes with experience. The same query issued at different times traverses different routes through the graph and recovers different context, since consolidated pathways carry more activation while decayed ones carry less or none. This is a structural departure from systems in which retrieval is a fixed function applied to a growing store. The effect is amplified by multi-cue convergence. All query entities seed simultaneously, each spreading its own activation wave through the graph, so retrieval naturally surfaces nodes where several independent signals converge through the association structure.

\section{Discussion} \label{sec:discussion}

Memini argues that as deployed LLM systems operate in a world that keeps shifting, the continual learning this requires can happen in the memory itself. Rather than a store managed from outside, memory becomes a substrate that evolves through coupled multi-timescale dynamics. These dynamics draw on synaptic consolidation models that resolve the same plasticity-stability tension found in biological memory. From this single mechanism, episodic sensitivity, gradual consolidation, and selective forgetting are expected to emerge together, without separate stores, explicit thresholds, or gating modules. What this opens up is a research direction in which memory is not the place where past information sits, but the place where the system learns.

Appendix~\ref{app:illustration} reports an initial check on a Wikipedia document stream, where the expected regimes emerge and the slow variable is shown to be necessary rather than incidental. Scaling this up to full empirical validation, including retrieval and benchmarks suited to memory that adapts to an evolving stream of evidence, is the immediate next step.

\section*{Acknowledgements}
This work is funded by the European Union Horizon MSCA DN programme FINALITY (G.A. 101168816).

\section*{Impact Statement}
This paper presents a theoretical design for continual knowledge updating in LLM systems through memory dynamics. We do not foresee specific negative societal consequences of this work that require further discussion here.

\bibliography{references}
\bibliographystyle{icml2026}

\newpage
\appendix
\onecolumn

\section{Testing the Dynamics on a Wikipedia Document Stream} \label{app:illustration}

This appendix supplements Section~\ref{sec:memory:emergent} by examining the coupled dynamics defined in Equations~\eqref{eq:wfast} and~\eqref{eq:wslow} on a stream of co-occurrence events extracted from real, temporally ordered text. The aim is twofold. First, we check that the three regimes derived analytically, namely episodic sensitivity, gradual consolidation, and selective forgetting, also emerge when the input is drawn from a real document stream. Second, we test whether the slow variable is necessary, by comparing Memini against a matched single-timescale ablation across all entity pairs in the stream. We use Wikipedia articles tracking the COVID-19 pandemic because the topic has a clear, widely understood phase structure, with associations that should plausibly consolidate within a phase such as \emph{vaccine} and \emph{mRNA} during the vaccine rollout, and others that should fade across phases such as \emph{bat} and \emph{SARS-CoV-2} once the discourse moves beyond origin. The pandemic's well-known timeline therefore provides an external check on whether the dynamics behave sensibly.

\subsection{Document Stream and Versioning}

The document stream consists of 13 English Wikipedia articles on COVID-19 topics, ordered by the period in which each topic became prominent during the pandemic. It spans four phases, namely origin (Phase~1, step~0), containment (Phase~2, steps~1--4), vaccines (Phase~3, steps~5--7), and variants and endemic transition (Phase~4, steps~8--12).

Wikipedia articles are continuously edited, so using current revisions would contaminate early time steps with content written years after the events they describe. To prevent this, every article is fetched through the MediaWiki revision API at the last edit on or before a target date corresponding to its phase, ensuring that every sentence used for co-occurrence extraction existed in Wikipedia at that date. Table~\ref{tab:articles} lists the resulting articles, revision identifiers, and timestamps, and any reported event can be audited by visiting the corresponding revision on Wikipedia.

\begin{table}[h]
\centering
\caption{Wikipedia articles used as the document stream, with the revision identifier and timestamp at which each article was fetched. Each article corresponds to one time step in the input sequence. Each revision identifier links to the exact historical revision used for content extraction, allowing any reported event to be independently verified.}
\label{tab:articles}
\footnotesize
\begin{tabular}{clccc}
\toprule
\textbf{Step} & \textbf{Article} & \textbf{Phase} & \textbf{Revision ID} & \textbf{Revision date} \\
\midrule
0  & SARS-CoV-2                              & 1 & \href{https://en.wikipedia.org/w/index.php?oldid=943272842}{943272842}   & 2020-02-29 \\
1  & COVID-19 lockdowns                      & 2 & \href{https://en.wikipedia.org/w/index.php?oldid=954085088}{954085088}   & 2020-04-30 \\
2  & Face masks during the COVID-19 pandemic & 2 & \href{https://en.wikipedia.org/w/index.php?oldid=960065515}{960065515}   & 2020-05-31 \\
3  & COVID-19 testing                        & 2 & \href{https://en.wikipedia.org/w/index.php?oldid=965344908}{965344908}   & 2020-06-30 \\
4  & Hydroxychloroquine                      & 2 & \href{https://en.wikipedia.org/w/index.php?oldid=970150834}{970150834}   & 2020-07-29 \\
5  & COVID-19 vaccine                        & 3 & \href{https://en.wikipedia.org/w/index.php?oldid=991568447}{991568447}   & 2020-11-30 \\
6  & Moderna COVID-19 vaccine                & 3 & \href{https://en.wikipedia.org/w/index.php?oldid=1003792446}{1003792446} & 2021-01-30 \\
7  & mRNA vaccine                            & 3 & \href{https://en.wikipedia.org/w/index.php?oldid=1009464972}{1009464972} & 2021-02-28 \\
8  & SARS-CoV-2 Delta variant                & 4 & \href{https://en.wikipedia.org/w/index.php?oldid=1036472511}{1036472511} & 2021-07-31 \\
9  & SARS-CoV-2 Omicron variant              & 4 & \href{https://en.wikipedia.org/w/index.php?oldid=1058010984}{1058010984} & 2021-11-30 \\
10 & Booster dose                            & 4 & \href{https://en.wikipedia.org/w/index.php?oldid=1062765134}{1062765134} & 2021-12-30 \\
11 & Long COVID                              & 4 & \href{https://en.wikipedia.org/w/index.php?oldid=1095797409}{1095797409} & 2022-06-30 \\
12 & Endemic COVID-19                        & 4 & \href{https://en.wikipedia.org/w/index.php?oldid=1130409555}{1130409555} & 2022-12-30 \\
\bottomrule
\end{tabular}
\end{table}

\subsection{Extracting the Event Stream}

We track 20 entities spanning all four pandemic phases, with five entities per phase covering origin, containment, vaccines, and variants. These entities act as the nodes of the association graph. They are detected in text by case-insensitive word-boundary string matching against the 20 predefined terms and a small set of hand-curated aliases, with abbreviations such as WHO and PCR matched case-sensitively to avoid spurious hits. Co-occurrence is detected at the sentence level, with each entity pair generating at most one event per document regardless of how many sentences contain both. This choice favors stronger signals of semantic association and makes the temporal pattern across documents, rather than within them, the driver of consolidation. Across the 13 documents, the procedure produces 124 events covering 68 unique entity pairs.

Each pair is integrated independently using the same parameters. We use the simplest two-variable instantiation of the Benna-Fusi chain ($n=2$), with $\tau_{\text{fast}}=2$, $\tau_{\text{slow}}=10$, coupling $C=0.2$, binary input $b=1$, and a forward Euler step $\Delta t = 1$ corresponding to one document per step. Both variables are clamped to be non-negative. The timescale ratio $\tau_{\text{slow}}/\tau_{\text{fast}}=5$ is chosen so that the three expected regimes are clearly distinguishable within the 13-step window. The remaining parameters are not tuned per pair and are held fixed across all events.

\subsection{Visualizing the Dynamics across Regimes}

\begin{figure}[h]
    \centering
    \includegraphics[width=0.95\linewidth]{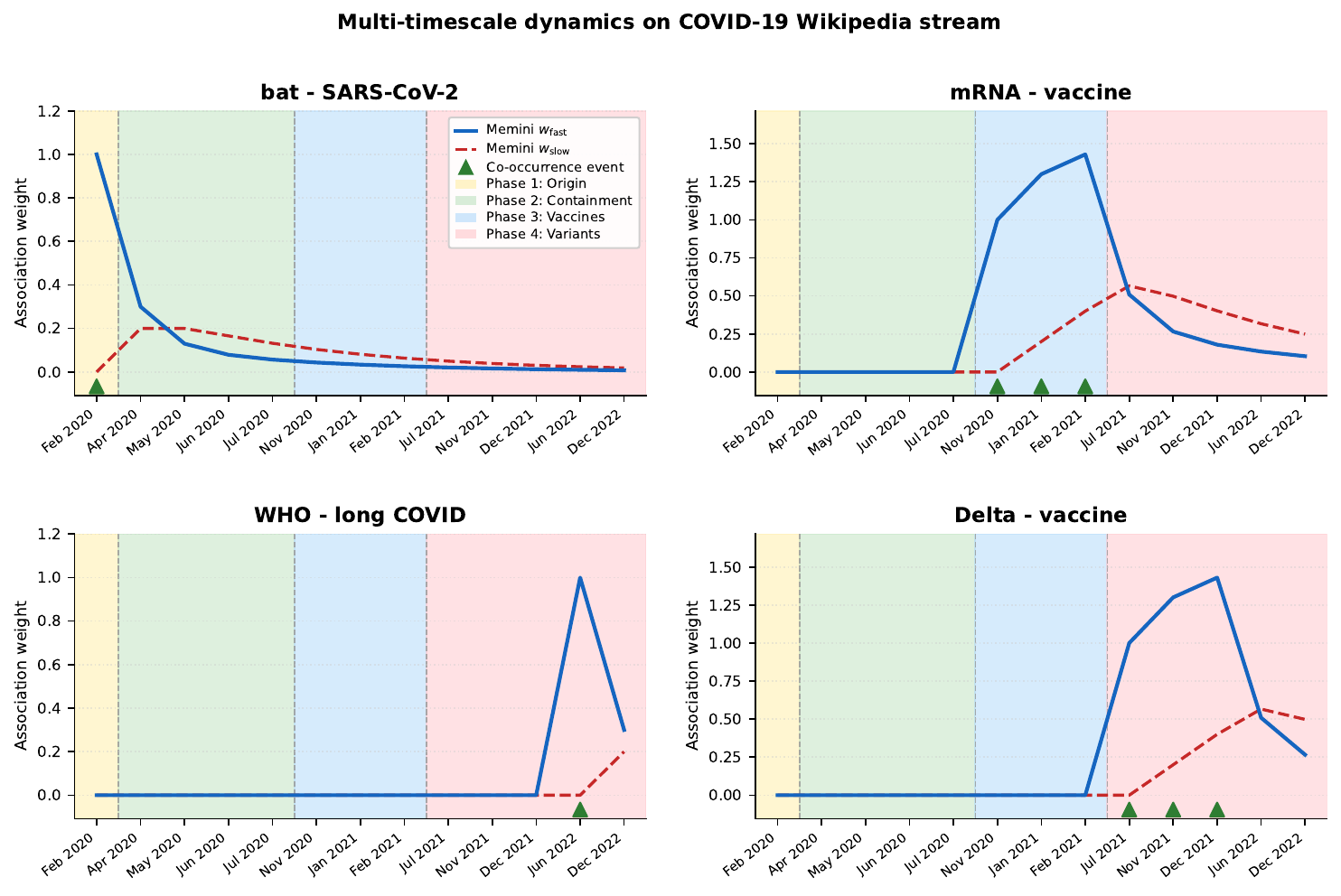}
    \caption{Multi-timescale association dynamics on the COVID-19 Wikipedia stream. Solid blue: $w_{\text{fast}}$. Dashed red: $w_{\text{slow}}$. Green triangles: co-occurrence events, each corresponding to a specific sentence in a versioned revision. Background colors indicate pandemic phase. Each panel illustrates one of the regimes analyzed in Table~\ref{tab:ablation}.}
    \label{fig:app_dynamics}
\end{figure}

Figure~\ref{fig:app_dynamics} shows the $w_{\text{fast}}$ and $w_{\text{slow}}$ trajectories for four entity pairs across the 13-document stream under identical parameters. The four were selected so that each illustrates one of the event-pattern groups later defined in Table~\ref{tab:ablation}, with all differences between panels arising entirely from the temporal pattern of co-occurrence events.

The panels illustrate the three regimes from Section~\ref{sec:memory:emergent}. \emph{Bat -- SARS-CoV-2} appears once at step~0 in the description of the virus's suspected origin in bats and never again. This is the episodic regime, where $w_{\text{fast}}$ spikes and decays within a few steps while $w_{\text{slow}}$ rises briefly through coupling but never accumulates. \emph{mRNA -- vaccine} appears in three consecutive Phase~3 articles tracking the vaccine rollout. The clustered events drive $w_{\text{slow}}$ upward, and the elevated $w_{\text{slow}}$ then sustains $w_{\text{fast}}$ above the episodic baseline well into 2022 after direct input stops, illustrating consolidation. \emph{WHO -- long COVID} appears once near the end of the stream, reflecting the late institutional formalization of long COVID as a recognized condition, and produces a sharp $w_{\text{fast}}$ response with no accumulated $w_{\text{slow}}$ to sustain it. \emph{Delta -- vaccine} appears in three consecutive Phase~4 articles on Delta, Omicron, and booster doses, showing ongoing consolidation, with $w_{\text{slow}}$ still rising near the end of the stream because the events occurred too recently to have decayed.

Recency itself emerges from the dynamics rather than being prescribed. The \emph{mRNA -- vaccine} and \emph{Delta -- vaccine} pairs receive the same number of events with the same spacing, yet \emph{Delta -- vaccine} retains a higher final state because its events arrive later in the stream and have had less time to decay. This recency sensitivity follows directly from Equations~\eqref{eq:wfast} and~\eqref{eq:wslow}.

\subsection{Is the Slow Variable Necessary? An Ablation across All Pairs}

Figure~\ref{fig:app_dynamics} illustrates that the dynamics behave qualitatively as expected on a small set of representative pairs, but it cannot establish that the slow variable is necessary. To test this, we compare Memini against two alternatives on the full set of 68 entity pairs in the stream. The first is a single-timescale ablation in which $w_{\text{slow}}$ is removed and the remaining variable evolves as $dw/dt = -w/\tau + I$. We choose $\tau$ to match Memini's effective decay rate when $w_{\text{slow}}=0$, namely $\tau = 1/(1/\tau_{\text{fast}} + C) \approx 1.43$, so that the two systems are mathematically identical in the absence of consolidation. Any divergence between them must therefore originate from $w_{\text{slow}}$. The second is a uniform-retention baseline that simply accumulates event counts without any decay, representing a memory that never forgets.

We classify each of the 68 pairs into one of four groups using only two quantities. The first is the number of co-occurrence events $n$ that the pair receives across the stream. The second is the document index at which the most recent of those events occurs, where documents are indexed 0 to 12, so a higher index means more recent. \emph{Repeated, no longer mentioned} pairs ($n \geq 3$, last index $\leq 7$) are associations that should require multi-timescale dynamics to be retained. \emph{Few mentions, no longer mentioned} pairs ($n \in \{1,2\}$, last index $\leq 7$) are weakly supported associations that should be forgotten. \emph{Repeated, recently mentioned} pairs ($n \geq 3$, last index $\geq 10$) are strongly supported and current. \emph{Few mentions, recently mentioned} pairs ($n \in \{1,2\}$, last index $\geq 10$) are weakly supported but current. Pairs whose last event falls at index 8 or 9 are excluded so that the old and recent regions remain cleanly separated. The four-way classification covers 51 of the 68 pairs and is determined entirely by event metadata.

Table~\ref{tab:ablation} reports the mean final association weight in each group under each of the three configurations, and the four rows together produce the expected pattern. The headline result is in the first row. Repeated associations that are no longer mentioned are exactly the case where multi-timescale architecture should matter most, and Memini retains a mean weight roughly thirty times larger than the single-timescale ablation. This gap cannot be explained by faster decay, since the two systems share the same effective decay rate when $w_{\text{slow}}=0$, and it isolates the contribution of the slow variable. The remaining three rows act as controls. Few-mention associations that are no longer mentioned collapse to near zero in both Memini and the ablation, while uniform retention still carries them at a mean of 1.28, showing the cost of having no decay at all. Repeated associations that remain recently mentioned look similar across Memini and the ablation, since recent input dominates and consolidation has had less time to make its effect visible. Few-mention associations that are recently mentioned are tied within numerical precision, confirming that Memini gains no spurious advantage where consolidation has not had time to occur.

\begin{table}[h]
\centering
\caption{Mean final association weight across the 51 entity pairs classified by event pattern, with $N$ denoting the number of pairs in each group. Memini denotes the full coupled dynamics. The single-timescale ablation uses $\tau$ matched to Memini's effective early-decay rate when $w_{\text{slow}}=0$, so any divergence between Memini and the ablation is attributable to $w_{\text{slow}}$. Uniform denotes cumulative event count with no decay. The pattern labels follow the conventions defined in the main text, where ``repeated'' means three or more events, ``few mentions'' means one or two, ``no longer mentioned'' means the last event occurred at least five documents before the end of the stream, and ``recently mentioned'' means the last event occurred in the final three documents. Parameters are $\tau_{\text{fast}}=2$, $\tau_{\text{slow}}=10$, and $C=0.2$.}
\label{tab:ablation}
\small
\begin{tabular}{lcccc}
\toprule
Pattern & $N$ & Memini $w_{\text{fast}}$ & Single-timescale & Uniform \\
\midrule
Repeated, no longer mentioned     & 5  & 0.104 & 0.003 & 3.000 \\
Few mentions, no longer mentioned & 25 & 0.026 & 0.000 & 1.280 \\
Repeated, recently mentioned      & 8  & 0.426 & 0.260 & 4.625 \\
Few mentions, recently mentioned  & 13 & 0.463 & 0.451 & 1.000 \\
\bottomrule
\end{tabular}
\end{table}

The pattern across all four rows supports a sharper claim than any single comparison. Memini retains old associations only when those associations were repeatedly confirmed, fades old associations that received sparse support, and treats recent associations consistently with whatever evidence has accumulated for them. Selective retention here is not produced by an explicit deletion rule, a threshold, or a gating module, but emerges from the coupled dynamics applied uniformly to every pair. This positions forgetting not as a failure mode to be avoided, but as a functional property of the substrate, in line with views of memory in which selective decay supports adaptation rather than degrading it \cite{norbyWhyForgetAdaptive2015}. In Memini, the same mechanism that allows associations to persist also allows them to fade, and the architecture itself decides which.

\subsection{Scope}

This experiment is intentionally limited in scope. The corpus contains 13 documents, the four classified groups range from 5 to 25 pairs, and we report no retrieval metrics or comparison against published systems. What carries the result is therefore not any single group but the consistent pattern across all four, with Memini diverging from the matched ablation only where expected. The broader empirical questions, including validation on larger and more varied document streams, retrieval quality on temporal question-answering corpora, and comparison against existing memory-augmented systems, remain the appropriate targets for rigorous evaluation and are left to future work, as outlined in Section~\ref{sec:discussion}.

\end{document}